\documentclass{article}

\usepackage{skyreels_a1_arxiv}

\usepackage[utf8]{inputenc} 
\usepackage[T1]{fontenc}    
\usepackage{hyperref}       
\usepackage{url}            
\usepackage{booktabs}       
\usepackage{amsfonts}       
\usepackage{nicefrac}       
\usepackage{microtype}      
\usepackage{xcolor}         
\usepackage{algorithm}
\usepackage{epsfig}
\usepackage{amsmath}
\usepackage{amssymb}
\usepackage{booktabs}
\usepackage{bm}
\usepackage{multirow}
\usepackage{algorithmicx}
\usepackage{algpseudocode}
\usepackage{makecell}
\usepackage{textcomp}
\usepackage{bbding}
\usepackage{microtype}
\usepackage{array}
\usepackage{setspace}
\usepackage{threeparttable}
\usepackage{diagbox}
\usepackage{wrapfig}
\usepackage{float}
\usepackage{bbm}
\usepackage{url}
\usepackage{mathdots}

\title{SkyReels-A1: Expressive Portrait Animation in \\ Video Diffusion Transformers}

\author{%
  David S.~Hippocampus\thanks{Use footnote for providing further information
    funding agencies.} \\
  Department of Computer Science\\
  Cranberry-Lemon University\\
  Pittsburgh, PA 15213 \\
  \texttt{hippo@cs.cranberry-lemon.edu} \\
}

\begin{document}
\maketitle

\vspace{-0.5cm}
\begin{figure}[H]
  \includegraphics[width=\textwidth]{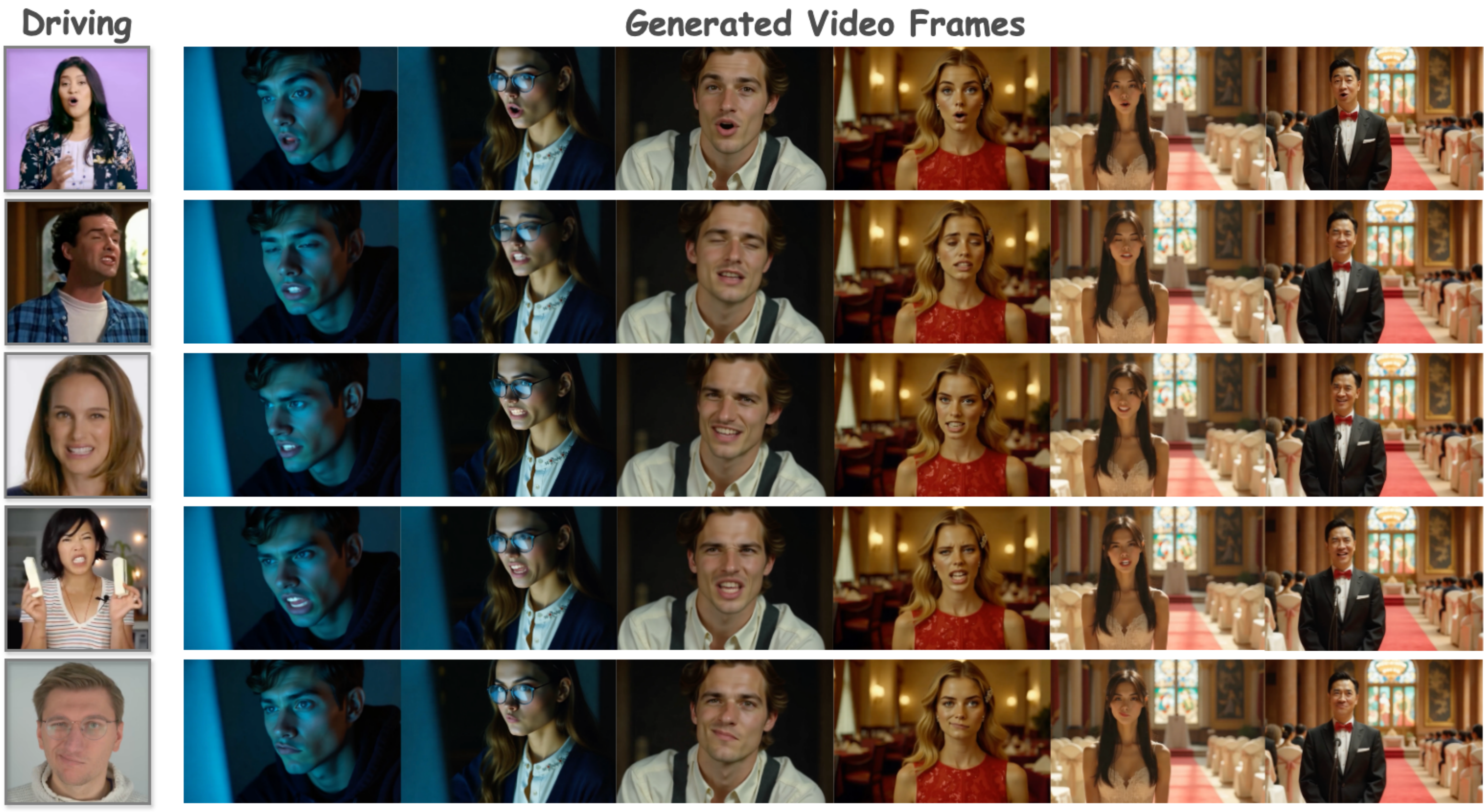}
  \caption{\textbf{SkyReels-A1 can generate an animated portrait from a reference image, driven by video motions while ensuring identity preservation.} Notably, our method ensures accurate transfer of facial expressions and body movements, allowing for realistic, high-quality portrait animations at anybody proportion that integrate naturally into different scenes.}
  \label{Fig: fig1}
\end{figure}

\begin{abstract}
We present \text{SkyReels-A1}\footnote{Core contributers: \texttt{\{qiudihk, feizhengcong\}}@gmail.com.}, a simple yet effective framework built upon video diffusion Transformers (DiT) to facilitate portrait image animation. 
Existing methodologies still encounter issues, including identity distortion, background instability, and unrealistic facial dynamics, particularly in head-only animation scenarios. Besides, extending to accommodate diverse body proportions usually leads to visual inconsistencies or unnatural articulations.
To address these challenges, SkyReels-A1 capitalizes on the strong generative capabilities of video DiT, enhancing facial motion transfer precision, identity retention, and temporal coherence. The system incorporates an expression-aware conditioning module that enables seamless video synthesis driven by expression-guided landmark inputs. Integrating the facial image-text alignment module strengthens the fusion of facial attributes with motion trajectories, reinforcing identity preservation.
Additionally, SkyReels-A1 incorporates a multi-stage training paradigm to incrementally refine the correlation between expressions and motion while ensuring stable identity reproduction. Extensive empirical evaluations highlight the model’s ability to produce visually coherent and compositionally diverse results, making it highly applicable to domains such as virtual avatars, remote communication, and digital media generation. The code and model weights of SkyReels-A1 are publicly available at \url{https://github.com/SkyworkAI/SkyReels-A1}.

\end{abstract}

\section{Introduction}
Generation of dynamic facial movements and expressions from a single static portrait, known as portrait animation, has emerged as a prominent research area due to its fascinating and diverse applications, including virtual avatars, digital communication, and narrative-driven media. Recent progress, especially with diffusion-based techniques\cite{ho2020denoising,song2020denoising,song2020score}, has significantly advanced the domain. Earlier approaches \cite{siarohin2019animating,siarohin2019first,siarohin2021motion,wang2021facevid2vid,1020894,megaportraits,Siarohin_2019_NeurIPS} predominantly relied on generative adversarial networks (GANs) \cite{goodfellow2020generative} to animate head movements, employing facial landmark tracking or 3D morphable models \cite{ye20203d} as motion guidance mechanisms. More recent developments\cite{hu2023animateanyone, xu2023magicanimate, chang2024magicpose} have integrated diffusion models to improve temporal stability and realism, with large pre-trained video generative models \cite{xie2024x,ma2024follow,chen2024echomimic,fei2024scaling,fei2024scalable}. It demonstrates superior capabilities in capturing intricate facial dynamics\cite{blattmann2023stable,yang2024cogvideox,fei2024video}, and enhances both pose-driven facial synthesis and audio-driven talking head generation. Moreover, commercial applications such as Runway’s Act-One have achieved notable progress in producing expressive and dynamic animated portraits, achieving lifelike facial expressions while adapting performances to characters with varying proportions. 

Although it marks a substantial leap in the application of generative models for creative media production\cite{xie2024x,chen2024echomimic,ma2024follow,jiang2024loopy,wang2024v,tian2024emo}, producing high-quality animations that maintain natural expressiveness and identity consistency across diverse body compositions remains an unresolved challenge \cite{fei2025ingredients}, particularly when animation extends beyond head movements, i.e., diverse body proportions. Existing methods \cite{ma2024follow, guo2024liveportrait} are constrained by issues such as identity drift, unstable background generation, and overly rigid facial expressions, which become more pronounced in full-body animation scenarios \cite{gowda2023pixels}. These challenges largely arise from the difficulty of seamlessly integrating identity-preserving features with motion dynamics, the absence of fine-grained conditioning for nuanced expressions, and limitations in training methodologies for comprehensive motion adaptation.

This study introduces SkyReels-A1, an advanced framework built upon the video diffusion Transformers (DiT) architecture, aimed at producing high-fidelity and expressive portrait animations. Our approach is distinguished by three primary innovations: 
(\textbf{i}) We harness the representational capacity of DiT to jointly model intricate facial details and full-body dynamics within a unified latent space, effectively addressing challenges such as identity drift and background inconsistencies. 
(\textbf{ii}) We incorporate an expression-aware conditioning module, which fuses continuous landmark sequences with facial image-text alignment features, facilitating precise manipulation of subtle expressions, \emph{e.g.}, eyebrow elevation, and lip curvature while ensuring identity coherence. 
(\textbf{iii}) A multi-stage training paradigm is implemented, which iteratively enhances pose accuracy, identity stability, and motion realism in sequence, leading to more natural and lifelike animations.

By unifying these components, SkyReels-A1 achieves superior performance in generating portrait animations that maintain identity integrity, exhibit natural expressiveness, and seamlessly accommodate variations in body proportions. Extensive empirical evaluations indicate that our framework surpasses existing approaches in both quantitative assessments and user studies, particularly in handling complex anatomical structures and intricate micro-expressions. Additionally, the modular design of SkyReels-A1 facilitates straightforward integration into downstream applications, including real-time video editing systems and personalized avatar platforms. This work represents a significant advancement in portrait animation by reconciling identity preservation with motion fidelity, thereby offering a robust framework for next-generation digital human technologies.

In summary, our contributions are listed as:
\begin{itemize}
\item We present {SkyReels-A1}, a novel framework for portrait animation that employs a DiT architecture to enhance fidelity in motion transfer precision, identity retention, and temporal coherence. The framework incorporates a dynamic conditioning module informed by expression landmarks and a cross-modal alignment mechanism that bridges visual-textual semantic spaces.
\item A phased training methodology is formulated to incrementally refine motion-expression correlation and subject-specific feature invariance. Initial stages establish coarse motion-texture associations, while subsequent phases employ adversarial regularization to synchronize fine-grained facial dynamics with latent identity embeddings.
\item A comprehensive series of experiments has been conducted to evaluate the performance, which indicates that {SkyReels-A1}, yields highly effective outcomes, exhibiting seamless adaptability to a wide range of compositional variations. 
This versatility renders our method particularly well-suited for various applications, including but not limited to the creation of virtual avatars, enhancement of video conferencing experiences, and the production of digital content. 
Finally, to support further research and industry adoption, both the code and demonstration materials are publicly available.
\end{itemize}

\section{Related Works}

\subsection{GAN-based Portrait Animation}

Early approaches to portrait animation have utilized GANs for learning motion dynamics in a self-supervised manner. The seminal works \cite{Siarohin_2019_CVPR,Siarohin_2019_NeurIPS} typically involve a two-stage process: warping and rendering. These methods use sparse neural keypoints to predict head motion, which is then applied to warp the encoded identity features of the source. The warped features are subsequently passed through a generative decoder to produce animated frames, incorporating in-painted facial details and background elements. In subsequent research, several approaches have sought to improve the warping estimation, with notable contributions including the use of 3D neural keypoints \cite{wang2021facevid2vid}, depth information \cite{hong2022depth}, and thin-plate splines \cite{siarohin2021motion}. Additionally, ReenactArtFace \cite{10061279} employed a 3D morphable model to control expressions and poses, while ToonTalker \cite{gong2023toontalker} applied a transformer-based architecture to enhance the warping process, particularly in the context of cross-domain datasets. Concurrently, efforts to refine the rendering process have resulted in several advancements. For instance, MegaPortraits \cite{megaportraits} elevated the resolution of the generated portraits to megapixels by utilizing high-resolution image datasets. FADM \cite{1020894} proposed a coarse-to-fine animation framework, incorporating a diffusion process to enhance facial details from the initial outputs generated by FOMM \cite{Siarohin_2019_NeurIPS} and Face Vid2Vid \cite{wang2021facevid2vid}. Beyond traditional video reenactment, significant strides have been made in leveraging alternative driving signals, such as 3D facial priors \cite{deng2020disentangled,Khakhulin2022ROME,sun2023next3d,DECA:Siggraph2021,10203414} and audio cues \cite{guo2021adnerf,10.1145/3550469.3555393,fei2024flux}. Despite these advancements, existing methods primarily rely on explicit feature warping, with an emphasis on facial animation in talking scenarios. 


\subsection{Diffusion-based Portrait Animation}

Recent advancements in diffusion models \cite{ho2020denoising,song2020denoising,song2020score} have led to exceptional performance across a variety of generative tasks, including image \cite{saharia2022photorealistic}, video \cite{guo2023animatediff,blattmann2023stable,bao2023latentwarp}, and multi-view renderings \cite{liu2023zero1to3, liu2023one2345, gu2023diffportrait3d}. Latent diffusion models (LDMs) \cite{rombach2021highresolution,fei2024dimba} have further refined this approach by reducing computational costs, performing the diffusion process within a lower-dimensional latent space. In the context of portrait animation, pre-trained diffusion models \cite{saharia2022photorealistic, rombach2021highresolution} have been widely adopted for image-to-video tasks. Notably, several studies \cite{cao2023masactrl, lin2023consistent123} have demonstrated the effectiveness of integrating reference image features into the self-attention blocks of LDM UNets, enabling improved image editing and video generation while preserving the original appearance context. 
In addition,  ControlNet \cite{zhang2023adding} extends the LDM architecture by incorporating structural control signals, such as landmarks, segmentations, and dense poses, to guide controllable image generation. This has inspired several concurrent works \cite{hu2023animateanyone, xu2023magicanimate, chang2024magicpose,qiu2024moviecharacter,fang2024motioncharacter} that achieve state-of-the-art full-body portrait animation by seamlessly integrating appearance and motion controls along with temporal attention mechanisms \cite{guo2023animatediff, guo2023sparsectrl} within pre-trained UNet models. Despite these advances, the motion control signals employed in these methods—such as skeletons with or without facial landmarks \cite{hu2023animateanyone, chang2024magicpose} and dense poses \cite{xu2023magicanimate}—may not fully capture the original motion dynamics. 
Furthermore, recent approaches \cite{ma2024follow, chen2024echomimic} that focus on facial expression and motion control have introduced more expressive landmarks for enhanced facial representation. However, these methods face challenges when applied to diverse body proportions, as they struggle to maintain consistent facial dynamics when transitioning to half-body or full-body compositions. In this work, we investigate the use of an advanced video diffusion transformers architecture to enhance portrait animation capabilities.

\section{Methodology}
\subsection{Preliminary}
Diffusion models generate structured data by progressively reversing a noise corruption process, transforming random noise into coherent samples.  At each time step $t \in \{0, \ldots, T\}$, the model estimates the latent state $x_t$ conditioned on $x_{t+1}$, as described by:
\begin{equation}
    p_{\theta}(x_t|x_{t+1}) = \mathcal{N}(x_t;\widetilde{\mu}_t,\widetilde{\beta}_t \mathit{I}),
\end{equation}
where $\theta$ denotes the model parameters, $\widetilde{\mu}_t$ is the predicted mean, and $\widetilde{\beta}_t$ is the variance schedule.
The training objective typically utilizes a mean squared error (MSE) loss on the noise prediction $\hat{\epsilon}_\theta(x_t,t,c_\text{txt})$ as:
\begin{equation}
\mathcal{L}_\text{noise} = w \cdot \mathbb{E}_{t,c_\text{txt},\epsilon \sim \mathcal{N}(0,1)}\Big[\Vert 
    \epsilon - \epsilon_\theta(x_t,t,c_\text{txt})\Vert^2\Big]\text{,}
    \label{eq:diff_loss}
\end{equation}
where $c_\text{txt}$ denotes the text condition.
Transformer-based text-to-video diffusion models have shown significant promise in generating video content. Our proposed model, SkyReels-A1, leverages an advanced transformer-based latent diffusion framework, incorporating a 3D Variational Autoencoder (VAE) to map videos from the pixel domain into a latent space. Each fundamental transformer block within the model consists of a 3D spatial-temporal self-attention mechanism and a feed-forward network (FFN).  The text embedding $c_\text{txt}$ is obtained by the T5 encoder. 
Recent studies on controllable generation usually extend this framework by incorporating additional control signals, such as image condition $c_{\text{img}}$. It is achieved through a feature extractor $\tau_{\text{img}}$ that processes a reference image $r$: $c_{\text{img}} = \tau_{\text{img}}(r)$ and the noise prediction function in~\ref{eq:diff_loss} becomes $\epsilon_\theta(x_t,t,c_\text{txt}, c_{\text{img}})$.

\begin{figure*}[t]
\begin{center}
\includegraphics[width=0.95\linewidth]{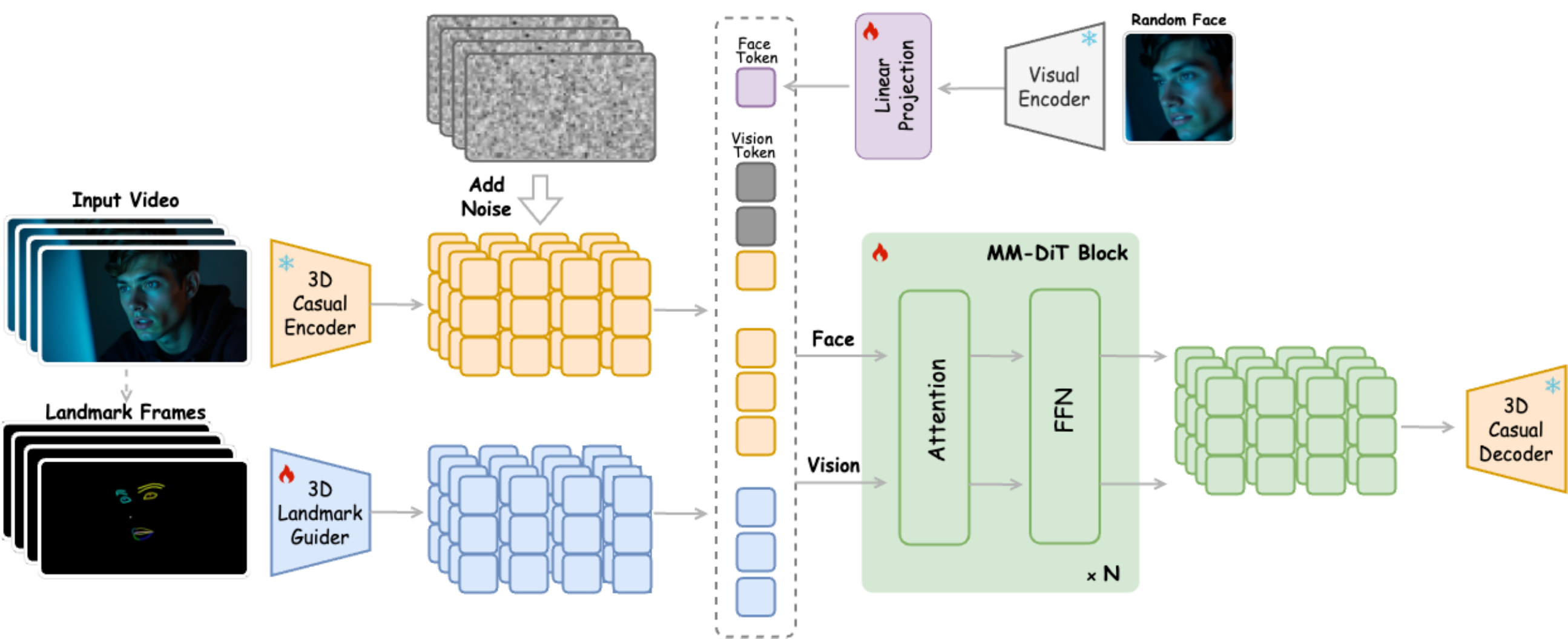}
\end{center}
   \caption{\textbf{Overview of SkyReels-A1 framework.} Given an input video sequence and a reference portrait image, we extract facial expression-aware landmarks from the video, which serve as motion descriptors for transferring expressions onto the portrait. Utilizing a conditional video generation framework based on DiT, our approach directly integrates these facial expression-aware landmarks into the input latent space. In alignment with prior research, we employ a pose guidance mechanism constructed within a VAE architecture. This component encodes facial expression-aware landmarks as conditional input for the DiT framework, thereby enabling the model to capture essential low-dimensional visual attributes while preserving the semantic integrity of facial features.}
\label{fig1}
\end{figure*}

\subsection{SkyReels-A1 Framework} 
\noindent \textbf{Overview.} 
The overall architecture of SkyReels-A1 is depicted in Figure \ref{fig1}. Given an input video clip and a reference portrait image $I_{ref}$, we extract facial expression-aware landmarks from the video as motion sequences to transfer the captured expressions onto the portrait. 
Based on a DiT-based conditional video generation framework, our model adopts a direct approach by incorporating the features of our facial expression-aware landmarks into the input latent. Similar to previous works, We employed a pose guider based on a VAE architecture to encode facial expression-aware landmarks as conditional input for DiT, ensuring that the model captures accurate low-dimensional visual information while preserving facial semantic information.
In between, the Facial Image-Text Alignment module enhances identity consistency by mapping facial features into the text feature space, enabling precise control over facial expressions and motion while preserving the character's identity. Specifically, we first use a face extractor $\mathcal{E}_{face}$ to locate the facial region $I_{face}$:
\begin{equation}
    I_{face} = \mathcal{E}_{face}(I_{ref}).
\end{equation}
To better extract the overall semantic information of the facial image $I_{face}$, facial features are obtained using a high-performance vision encoder $\mathcal{E}_{vision}$, i.e., SigLip\cite{zhai2023sigmoid}:
\begin{equation}
    F_{id} = \mathcal{E}_{vision}(I_{face}).
\end{equation}
Through the aforementioned strategies and architectural design, our method facilitates the generation of high-quality, expression-driven video content with improved realism and coherence.
In the following, we will describe the main module in SkyReels-A1 in detail.


\paragraph{Expression-Aware Landmark.} 
Inspired by \cite{ma2024follow, chen2024echomimic}, we introduce a similar facial landmark-based motion condition input for portrait animation. Accurate motion representation of facial expressions constitutes a fundamental requirement for expressive portrait animation, as it captures the nuanced spectrum of human emotions and micro-expressions that significantly enhance the realism and emotional resonance of animated avatars. Current diffusion-based approaches predominantly employ 2D facial landmarks as motion representations during training. However, both paradigms exhibit critical limitations: The reliance on 2D landmarks during inference frequently results in misalignment between target expressions and reference portraits, manifesting as expression mismatches and identity leakage artifacts. While alternative methods leverage third-party tools like MediaPipe to extract 3D keypoints from video sequences, such representations often lack the precision required to capture fine-grained expression details and complex facial dynamics, particularly in non-frontal orientations and extreme expressions. 

To overcome the aforementioned challenges, we propose 3D Facial Expressions\cite{retsinas20243d}, a framework that integrates a neural rendering module to enhance the accuracy and realism of reconstructed facial expressions. Unlike conventional approaches that rely on differentiable rendering, our method replaces this component with a neural rendering mechanism, facilitating a more effective learning process and improved generalization across diverse facial expressions. This architecture enables the extraction of high-precision 3D keypoints, capturing intricate motion details and complex facial dynamics with greater fidelity. By leveraging this refined motion representation, our approach significantly enhances the realism of portrait animations while ensuring improved expression accuracy, identity consistency, and adaptability across various application scenarios.

\paragraph{3D Landmark Guider.} 
To ensure spatio-temporal coherence between the driving signals and the input video latent representations, we propose the Spatio-temporal Alignment Landmark Guide Module. Specifically, it utilizes a 3D causal encoder as the core component for landmark guidance. Fine-tuned carefully, it can capture the driving signals' motion representation more effectively, ensuring precise and coherent alignment between the motion signals and the input video latents.

By leveraging the 3D causal encoder, the Spatio-temporal Alignment Landmark Guide module directly projects the driving signals into a shared latent space with the video latents. This shared representation bridges the gap between the motion signals and the generated video frames, resulting in synchronized spatial and temporal dynamics.
The fine-tuning process further refines the encoder’s ability to capture intricate motion patterns, thereby enhancing the realism and fidelity of motion transfer. This approach not only ensures precise motion alignment but also guarantees the preservation of both identity and motion consistency in the generated video, enabling high-quality, temporally stable animations.

\paragraph{Facial Image-Text Alignment.}
In existing portrait animation methods, maintaining consistent identity while altering facial expressions remains a challenging problem that requires further investigation. Specifically, earlier approaches have focused on enhancing identity consistency using identity-preserving adapters through cross-attention mechanisms. However, we found that this method is difficult to train and introduces a significant number of additional parameters. Inspired by the architecture of CogVideoX\cite{yang2024cogvideox}, we aim to improve facial identity consistency during the expression generation process by concatenating the embeddings of both facial images and video at the input stage. This approach not only enhances identity consistency but also allows for the seamless transfer of the pre-trained base model's capabilities. To achieve this, we introduce a lightweight learnable mapping module through a multi-layer perceptron $\mathcal{P}$, which maps facial features into the text feature space as:
\begin{equation}
    F'_{id} = \mathcal{P}(F_{id}),
\end{equation}
where $F_{id}$ is the identity embedding extracted by vision encoder $ \mathcal{E}_{vision}$ \cite{zhai2023sigmoid}, a pre-trained image-text model. Since $F_{id}$ captures detailed facial features while 
visual encompasses broader facial information and is less sensitive to external factors such as lighting and occlusion, the fusion of both types of information enhances the accuracy of facial features in the generated results.
The Face Image-Text Alignment Module facilitates better information integration while maintaining model efficiency.

\section{Training}

\subsection{Progressive Training Strategy} 
The training pipeline of SkyReels-A1 is structured into three distinct stages: Motion-Driven Training, Identity-Preserving Training, and Multi-module Joint Fine-Tuning. Each stage plays a pivotal role in progressively refining the model’s ability to generate high-fidelity, temporally consistent character animations.
\begin{itemize}
\item \textbf{Motion-Driven Training.} 
At this stage, we aim to incorporate motion conditions into the video generation process. Motion-driven video inputs are processed using a 3D landmark guidance module, which is initialized from a pre-trained 3D causal encoder but remains unchanged during this phase. To effectively integrate landmark-based motion information while maintaining the integrity of the image-to-video (IT2V) architecture, the extracted landmark latent representation is concatenated with the noise input. Model adaptation to landmark-specific variations is achieved by training only the convolutional layers within the PatchEmbedding module, ensuring that the system refines motion representation while preserving the core IT2V functionalities.
\item \textbf{Identity-Preserving Training.} 
In the subsequent stage, the focus shifts to reinforcing identity preservation in the animated portrait. Although textual conditions are not explicitly required for portrait animation, the text-processing branch is retained to capitalize on the pre-trained model’s capabilities. Facial features are encoded using a CLIP image encoder\cite{zhai2023sigmoid} to generate a facial representation, which is subsequently projected into the text feature space through a trainable linear mapping. Notably, only this projection layer undergoes optimization, while all other components remain unchanged. This design choice ensures that the animated portrait maintains identity consistency across varying expressions and motion sequences.
\item \textbf{Multi-module Joint Fine-Tuning.} This stage ensures the model's ability to generate precise, high-quality animations by jointly optimizing the 3D landmark guider, DiT block, and the linear projection layer. It further improves the model's generalization across diverse portrait types and motion scenarios, enabling it to handle a wide range of facial expressions and dynamic movements with greater fidelity.
\end{itemize}

\subsection{Face-Aware Loss}

The fluency of the generated video depends largely on the spatial coherence and realism of dynamic regions, such as the face. To enhance this, we apply a face-aware loss that directs the model to prioritize high-motion areas. 

Here we use RAFT~\cite{teed2020raft} as $\Theta$ for efficient and accurate optical flow estimation. Then the mean optical flow value $\tau_i$ can be calculated by simply averaging $f_{i,(x,y)}$.
Afterward, we take $\tau_i$ as the threshold to produce binary mask $\mathcal{M}_{i,(x,y)}$. Specifically, when the magnitude of the optical flow exceeds $\tau_i$, set the corresponding position in $\mathcal{M}_{i,(x,y)}$ to 1; otherwise set it to 0. Consequently, the mean foreground optical flow value $f_{i,fg}$ can be easily obtained by: 
\begin{equation}
    f_{i,fg} = \frac{1}{\mathcal{S}}\sum^H_{x=1}\sum^W_{y=1} f_{i,fg}(x,y) = \frac{1}{\mathcal{S}}\sum^H_{x=1}\sum^W_{y=1} \mathcal{M}_{i,(x,y)} * f_{i,(x,y)}, 
    \label{eq:foreground_optical_flow}
\end{equation}
where $f_{i,fg}(x,y)$ is the foreground optical flow at each pixel $(x, y)$. $\mathcal{S}$ denotes the number of the foreground pixels.
We then normalize the foreground optical flow $f_{i,fg}(x,y)$ defined in Equation \eqref{eq:foreground_optical_flow} and calculate the optical flow mask  $\mathcal{M}_{i,\text{norm}}$ as following: 
\begin{equation}
    \mathcal{M}_{i,\text{norm}} = \texttt{clip}\left(f_{i,fg}(x,y)/255 + 0.5, \, 1.0, \, 1.5 \right),
\end{equation}
where function $\texttt{clip}(\cdot, \, a, \, b)$ restricts the values into $[a, b]$. The high-motion areas will be assigned a greater value than the low-motion regions. 
Then the face-aware loss ${L}_{face}$ across all $N$ frames can be compactly defined as:
\begin{equation}
    {L}_{face} = \frac{1}{NH'W'} \sum_{i=1}^{N} \sum_{x=1}^{H'}\sum_{y=1}^{W'} \mathcal{M}_{i,\text{norm}} \cdot \left[ \epsilon_i(x,y) - \hat{\epsilon}_i(x,y) \right]^2, 
\end{equation}
where $\epsilon_i(x,y)$ and $\hat{\epsilon}_i(x,y)$ denote the target and predicted noise at location $ (x,y) $, respectively. $H'$ and $W'$ correspond to the resolution of latent features.





\begin{figure*}[t]
\begin{center}
\includegraphics[width=0.98\linewidth]{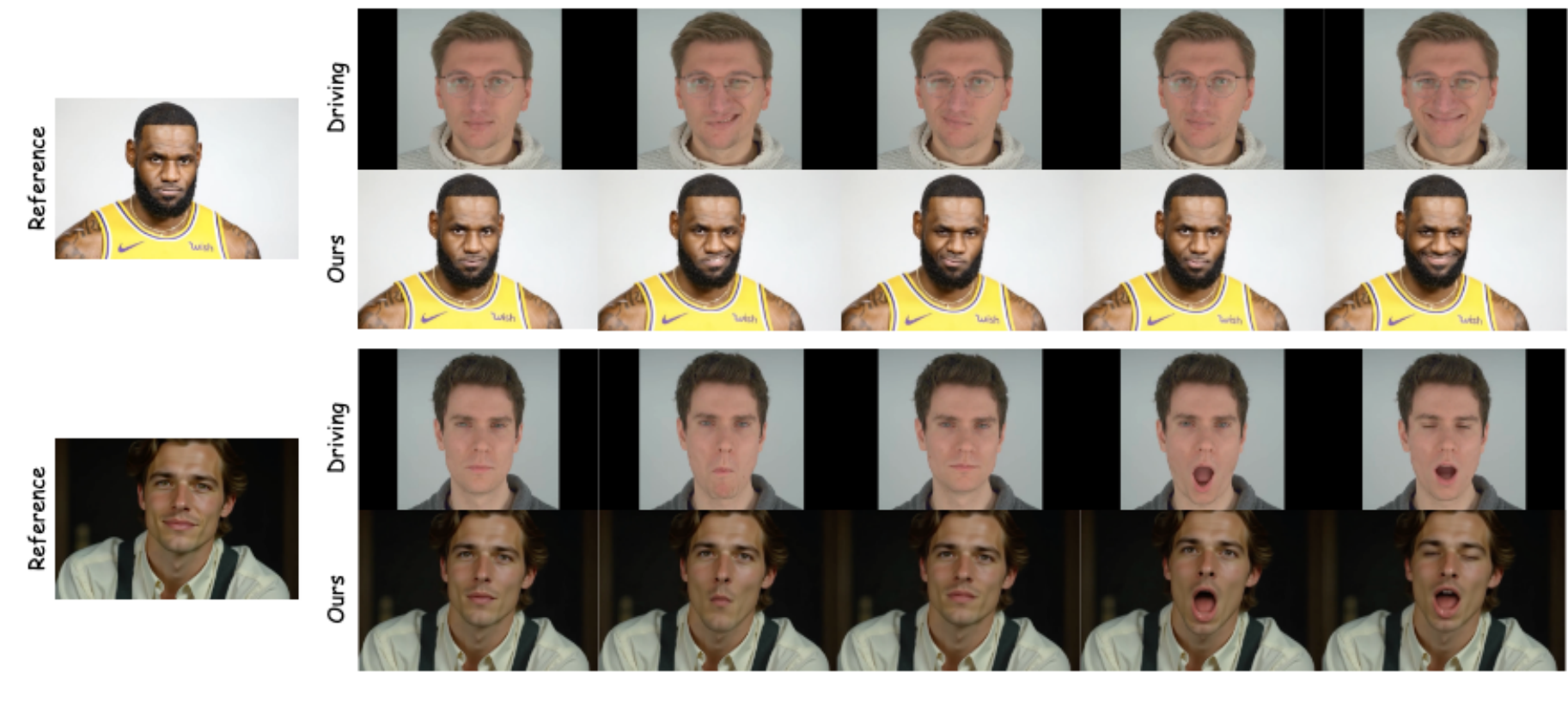}
\end{center}
   \caption{\textbf{Qualitative portrait animation results from SkyReels-A1.} Given a static portrait image as input, our model can vividly animate it, ensuring seamless stitching and offering precise control over eyes and lip movements. }
\label{fig2}
\end{figure*}

\section{Experiment}

We first give an overview of the implementation details, including data sources, data filtering processes, baselines, and benchmarks utilized in our experiments. Next, we present comparative experimental results against selected baselines, to validate the effectiveness of the proposed modules.

\subsection{Experimental Settings}
\paragraph{Implementation Details.}
We train our model based on the advanced video diffusion transformer CogVideoX-5B\cite{yang2024cogvideox} using a combination of our collected dataset and publicly available datasets.
During the multi-stage training phases, we train for 2K steps in the first stage, followed by another 2K steps in the second stage, and conclude with 1K steps in the final stage, using a batch size of 512. The learning rate is set to 1e-5 for the first two stages and reduced to 1e-6 in the final stage, utilizing AdamW optimizers. Our experiments are conducted on 32 NVIDIA A800 GPUs. During inference, we utilize DDIM sampler and set the scale of classifier-free guidance to 3 in our experiment. The static images used as reference are generated by Flux and sourced from the Pexels. 

\paragraph{Dataset Sources.} 
The training video clips featured are sourced from the NeRSemble\cite{kirschstein2023nersemble} dataset, HDTF\cite{zhang2021flow}, DFEW\cite{jiang2020dfew}, RAVDESS\cite{livingstone2018ryerson}, Panda70M\cite{chen2024panda} and our
collected dataset that we collected approximately 1W character video clips from the internet.

\paragraph{Dataset Filtering.} 
During the data pre-processing stage, we implement a series of meticulous filtering steps to ensure the quality and suitability of the video-text dataset. The workflow consists of three phases: single-character extraction, motion filtering, and post-processing. First, we select single-character videos to clean the video content using existing tools, addressing issues such as camera capture artifacts and background noise. Next, we extract head pose information and mouth landmarks using facial key points detected by MediaPipe. By calculating head angles and mouth variations, we filter samples with significant facial expressions and head movements.
Finally, based on the detected facial positions from the previous steps, We crop or pad the videos to a fixed resolution of 480$\times$720 to meet the model’s input requirements. A random frame is then selected from each video, and the face is encoded into an embedding using clip encoder\cite{zhai2023sigmoid}, providing essential facial feature information for the model.


\begin{table*}[t]
    \centering
    \setlength{\tabcolsep}{2.8mm}{
      \begin{tabular}{lcccc}
        \toprule 
        Method & \!\!ID Similarity$\uparrow$\!\! & \!\!Image Quality$\downarrow$\!\! & \!\!Expression Dis.$\downarrow$\!\! & \!\!Pose Dis.$\downarrow$\!\! \\
        \midrule
        Follow-Your-Emoji \cite{ma2024follow}  &0.5771 / 0.5983 & 100.2192 &0.0417 & 0.8502\\
          LivePortrait \cite{guo2024liveportrait} & 0.7011 / 0.7305 & 83.3168 & 0.0396 & 0.8372 \\
        Act-One & 0.7219 / 0.7470 & 68.5953  &0.0329 & 0.8204 \\
        \midrule
        \textbf{SkyReels-A1} & 0.7196 / 0.7314 &\textbf{59.6884} & 0.0363 & 0.8245\\
         \bottomrule 
    \end{tabular}
    }
    \caption{\textbf{Quantitative comparisons of portrait animation.} SkyReels-A1 exhibits a substantial enhancement in image quality, outperforming all existing approaches by a notable margin, with the exception of the proprietary commercial model Act-One.}
    \label{tab:comparison}
\end{table*}

\begin{figure*}[t]
\begin{center}
\includegraphics[width=0.98\linewidth]{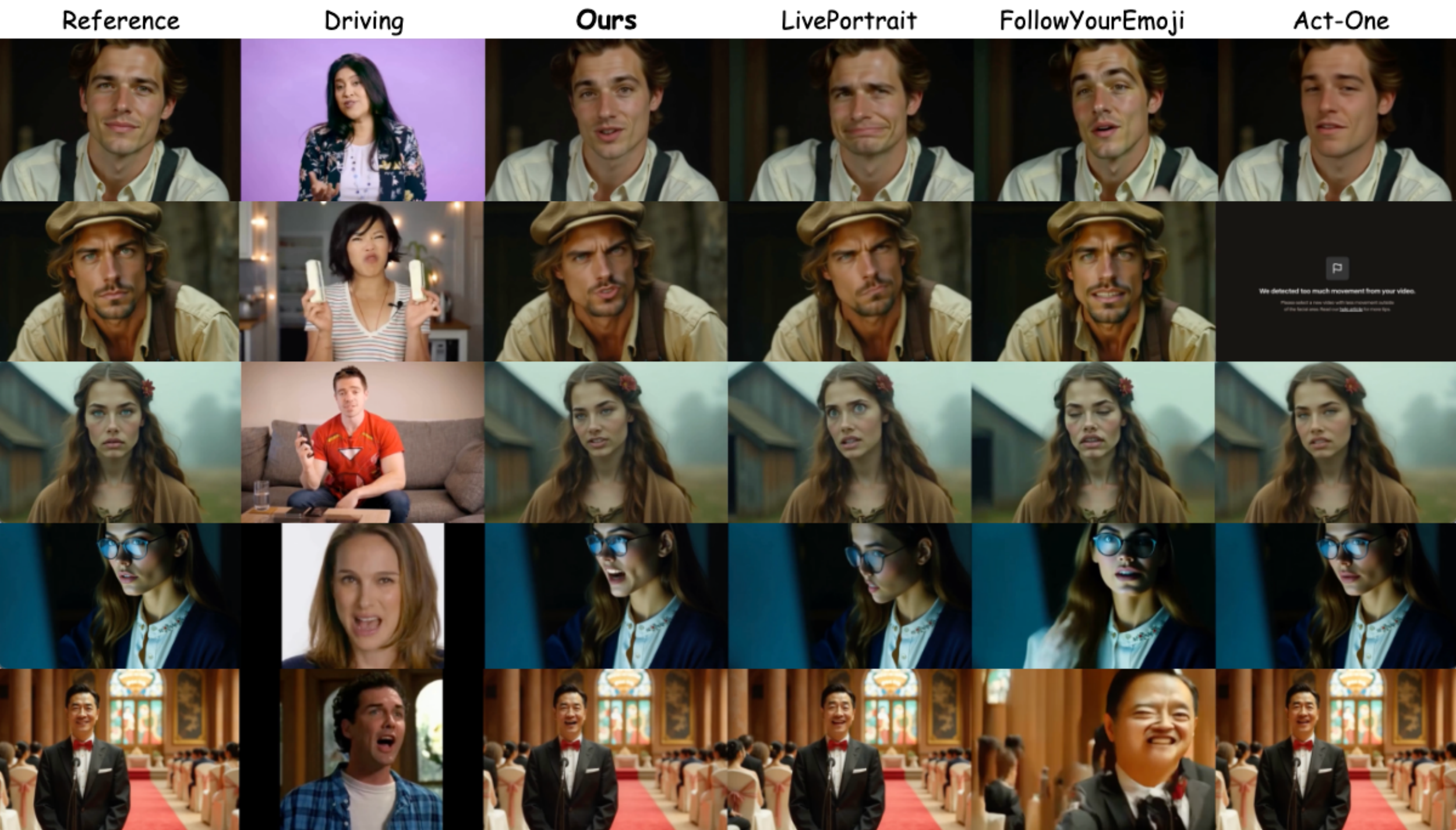}
\end{center}
   \caption{\textbf{Qualitative comparisons results.} Our SkyReels-A1 model better transfers lip movements and eye gazes from another person, while maintaining the identity of the source
portrait.}
\label{fig4}
\end{figure*}

\begin{figure*}[t]
\begin{center}
\includegraphics[width=0.94\linewidth]{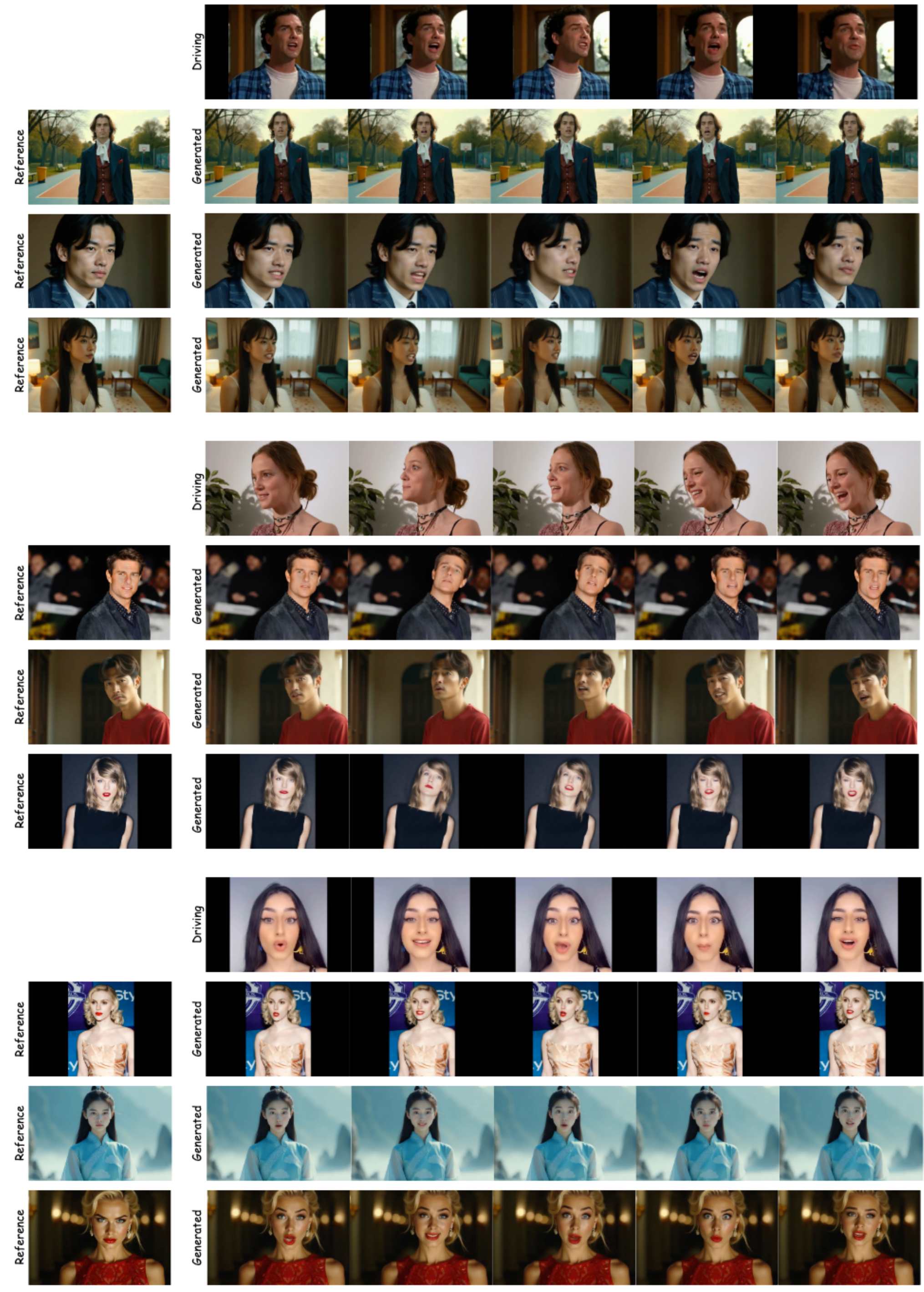}
\end{center}
   \caption{\textbf{More generated results from Skyreels-A1 in diverse body proportions.}}
\label{fig5}
\end{figure*} 

\paragraph{Baselines.} To comprehensively assess SkyReels-A1's performance across diverse scenarios, we compare it with portrait animation baselines, including the open-source solution LivePortrait\cite{guo2024liveportrait}, Follow-Your-Emoji\cite{ma2024follow}, and the closed-source commercial product Runway Act One\footnote{https://runwayml.com/research/introducing-act-one}.

\paragraph{Evaluation Metrics.} 
To measure the generalization quality and motion accuracy of portrait animation results, we employ three metrics to evaluate identity similarity, image quality, and expression and head pose accuracy, respectively. Specifically, the FaceSim-Arc and FaceSim-Cur score~\cite{deng2019arcface}, calculating the cosine similarity between source and generated images, is utilized to assess identity preservation. We employ a pre-trained network with FID~\cite{heusel2017gans} for image quality assessment. To evaluate the motion accuracy, we compare the L1 difference between the extracted facial blendshapes and head poses of the driving and generated frames using FaceAnalysis\footnote{https://github.com/deepinsight/insightface} and OpenPose\footnote{https://github.com/CMU-Perceptual-Computing-Lab/openpose}. 

\subsection{Comparison with Baselines}

\paragraph{Quantitative Results.}
We conducted experiments to investigate cross-identity motion transfer, wherein a reference portrait was randomly selected from a set of 100 in-the-wild images, while the driving sequence was obtained from our test dataset.
Table \ref{tab:comparison} presents the quantitative evaluation results. We can see that our proposed model demonstrates superior performance compared to both diffusion-based and non-diffusion-based approaches in terms of generation fidelity and motion precision. Notably, by incorporating video diffusion transformers as a prior, SkyReels-A1 achieves a significant improvement in image quality, surpassing existing methods by a considerable margin except the closed-source commercial model Act-One.

\paragraph{Qualitative Results.}
Figure \ref{fig4} presents qualitative comparisons of portrait animation, complementing the results obtained from automated evaluation metrics. The first two examples highlight the model’s robustness in transferring motion accurately, even under significant pose variations in the driving or source portraits. In the third and fourth cases, the model effectively captures and transfers subtle facial expressions, such as lip movements and eye gazes, while preserving the visual consistency of the original portrait. Furthermore, the final case demonstrates that the integration of stitching techniques enhances the model’s stability, allowing it to animate full-body images seamlessly, even when the reference portrait contains a relatively small facial region.

\subsection{User Study} 
To further validate the superiority of the proposed SkyReels-A1 model over prior baselines in motion accuracy and expressiveness, a user study was conducted. Specifically, 20 participants from diverse geographic regions were recruited to evaluate synthesized videos. Each participant responded to a series of comparative questions assessing two key aspects: motion accuracy and human likeness. The evaluation was conducted with explicit knowledge of the model names, and participants were instructed to select the video that most accurately replicated the expressions and movements of the driving sequence. Among the 100 collected responses, 63\% indicated a preference for SkyReels-A1, affirming its enhanced capability in preserving both facial expressions and pose fidelity compared to existing baselines.



\section{Conclusion}

In this study, we present SkyReels-A1, an innovative portrait animation framework based on video diffusion transformers. By integrating motion and identity representations, our approach achieves high-fidelity generation of both subtle and exaggerated facial expressions. Through extensive automatic and user-based evaluations, we have demonstrated the robustness and adaptability of our model across various customized scenarios. We anticipate that these promising results will contribute to advancements in portrait animation applications.

\paragraph{Limitations.}
Despite its strong performance, the current model exhibits limitations when handling extreme pose variations. Addressing these challenges remains a key focus for future research.

\newpage



\bibliographystyle{ieee}
\bibliography{skyreels_a1_arxiv}

\end{document}